\RequirePackage{etex} 
\documentclass[runningheads,a4paper]{llncs}

\usepackage{etex}
\reserveinserts{28}
\pdfoutput=1
\usepackage[a4paper, total={7in, 10in}]{geometry}
\usepackage{ textcomp }
\usepackage{ comment }
\usepackage[color=yellow!60,textsize=footnotesize,obeyDraft,draft]{todonotes}
\usepackage{colortbl}
\usepackage{booktabs}
\usepackage{tabularx}
\usepackage{capt-of}
\usepackage{makecell}
\usepackage{multirow}
\usepackage{listings}
\usepackage{amsmath}
\usepackage{caption}
\usepackage{subcaption}
\usepackage[font=footnotesize,labelfont=bf]{subcaption}
\usepackage[hidelinks]{hyperref}
\usepackage[capitalize]{cleveref}
\usepackage{tikz}
\usepackage{pgfplots}
\pgfplotsset{compat=newest}
\pgfplotsset{plot coordinates/math parser=false}
\usepackage[para,online,flushleft]{threeparttable}
\usepackage{wasysym}
\usepackage{amssymb}
\usetikzlibrary{arrows}
\usepackage{breakcites}
\usepackage{array}
\usepackage{color}
\usepackage{balance} 
\usepackage{lastpage}
\usepackage{dblfloatfix}
\usepackage{authblk}
\usepackage{siunitx}
\usepackage{xurl} 
\usepackage[shortlabels]{enumitem}
\usepackage{soul}
\usepackage{xcolor,listings}
\usepackage{relsize}
\usepackage{algorithm}
\usepackage{algpseudocode}

\usepackage{float}
\floatstyle{plaintop}
\restylefloat{table}
\usepackage{tikz}
\usetikzlibrary{shapes.geometric, arrows,calc}
\tikzstyle{startstop} = [rectangle, rounded corners, minimum width=3cm, minimum height=1cm,text centered, draw=black, fill=none]
\tikzstyle{io} = [trapezium, trapezium left angle=70, trapezium right angle=110, minimum width=3cm, minimum height=1cm, text centered, draw=black, fill=blue!30]
\tikzstyle{process} = [rectangle, minimum width=3.5cm, minimum height=1cm, text centered, draw=black, fill=orange!0]
\tikzstyle{neuron} = [circle, minimum size=3cm, text centered, draw=black, fill=orange!0]
\tikzstyle{decision} = [diamond, minimum width=3cm, minimum height=1cm, text centered, draw=black, fill=green!30]
\tikzstyle{arrow} = [thick,->,>=stealth]

\usepackage[maxnames=6,firstinits=true,doi=false,url=true,isbn=false]{biblatex}
\begin{document} 

\title{Evaluation Metrics for DNNs Compression}
\titlerunning{Evaluation Metrics for CNNs Compression}
\authorrunning{Ghobrial et al.}

\author{Abanoub Ghobrial\inst{1}, Samuel Budgett\inst{3}, Dieter Balemans\inst{2}, Hamid Asgari\inst{3}, Phil Reiter\inst{2}, Kerstin Eder\inst{1}}

\institute{University of Bristol, Bristol, UK \and University of Antwerp, Antwerp, Belgium \and Thales, Reading, UK}
\tocauthor{Authors' Instructions}

\maketitle
\makeatletter
\renewcommand\subsubsection{\@startsection{subsubsection}{3}{\z@}%
                       {-18\p@ \@plus -4\p@ \@minus -4\p@}%
                       {4\p@ \@plus 2\p@ \@minus 2\p@}%
                       {\normalfont\normalsize\bfseries\boldmath
                        \rightskip=\z@ \@plus 8em\pretolerance=10000 }}
\makeatother

\textbf{\textit{Abstract}--There is a lot of ongoing research effort into developing different techniques for neural networks compression. However, the community lacks standardised evaluation metrics, which are key to identifying the most suitable compression technique for different applications. 
This paper reviews existing neural network compression evaluation metrics and implements them into a standardisation framework called NetZIP. 
We introduce two novel metrics to cover existing gaps of evaluation in the literature: 1) Compression and Hardware Agnostic Theoretical Speed (CHATS) and 2) Overall Compression Success (OCS).
We demonstrate the use of NetZIP using two case studies on two different hardware platforms (a PC and a Raspberry Pi 4) focusing on object classification and object detection.}

\section{Introduction}
State-of-the-art (SoTA) Deep Neural Networks (DNNs) are becoming the go-to solution in computing for automated operations in complex high dimensional domains. 
Achieving high accuracy DNN models during training usually comes at the expense of overparameterizing the model~\cite{Neill2020}.
This results in well-performing DNNs that are oversized. 
Being oversized hinders the deployment of DNN models on edge devices restricted by minimal resources, often resulting in having an unnecessary large carbon footprint during operation.
These issues are expected to get worse as the complexity of operational environments increase and, thus, the models will consequently need to be larger, making the deployment even more challenging~\cite{Marino2023, Brown2020}.  

A possible solution is to take an adaptive approach, where a smaller model is trained on a subset of the operational environment and then adapted depending on shifts in the environment e.g.~\cite{Ghobrial2022, mirza2022norm}.
These approaches present their own questions regarding information forgetting, resource requirements and energy consumption.
Also, whilst this may assist with the problem of needing to increase the model size as the operational environment increases,
the neural network will still be oversized compared to the amount of information retained at one time. 

Therefore, neural network compression has recently seen a surge in research interest, as compression can offer significant reductions in size and energy consumption whilst aiming to retain the same overall accuracy and improving the rate of predictions. 
%
There are many compression techniques in literature that fall under one of the categories: pruning, quantization, knowledge distillation, and tensor decomposition~\cite{Neill2020}\cite{Marino2023}.
%
While there is significant research focus on developing compression techniques, the community seems to lack a standardised way of evaluation.

Based on a recent review, most SoTA methods in the literature do not compare their compression techniques with different existing techniques, and where comparisons are made, they tend to vary between different publications~\cite{Blalock2020}. 
Quarter of the papers reviewed by Blalock et al.~\cite{Blalock2020} did not compare their compression method to other methods and half of the papers reviewed compared to a maximum of one other method from the literature. 
Furthermore they did not find consistency in the evaluation metrics and dataset/network pairs used in the literature. 
This is largely due to the lack of a standardised implementation of metrics, neural networks, and compression techniques. 
In this paper, we focus on the metrics aspect.
The contributions of this paper are three-fold: 
\begin{enumerate}
    \item Provide a review of existing evaluation metrics used in assessing and comparing compression techniques. 
    \item Introduce two novel metrics, (a) Compression and Hardware Agnostic Theoretical Speed (CHATS) and (b) Overall Compression Success (OCS), to fill identified gaps in evaluation metrics. 
    \item  Introduce NetZIP, an open-source library for deep neural network compression evaluation. The NetZIP library implements all of the evaluation metrics in the paper with some examples of usages (link to library: \url{https://github.com/TSL-UOB/NetZIP}).
    
\end{enumerate}

The paper is organised as follows, in section~\ref{sec:background} we overview compression methods and existing SoTA compression benches. We provide our review of current evaluation metrics and our novel metrics in section~\ref{sec:metrics}. Section~\ref{sec:NetZIP_overview} introduces NetZIP and section~\ref{sec:casestudies} provides case studies to showcase comparisons made using NetZIP. Section~\ref{sec:results} discusses results. Conclusions and futures work are provided in section~\ref{sec:Conclusions}.

\begin{figure}
    \centering
    \includegraphics[width=1\textwidth]{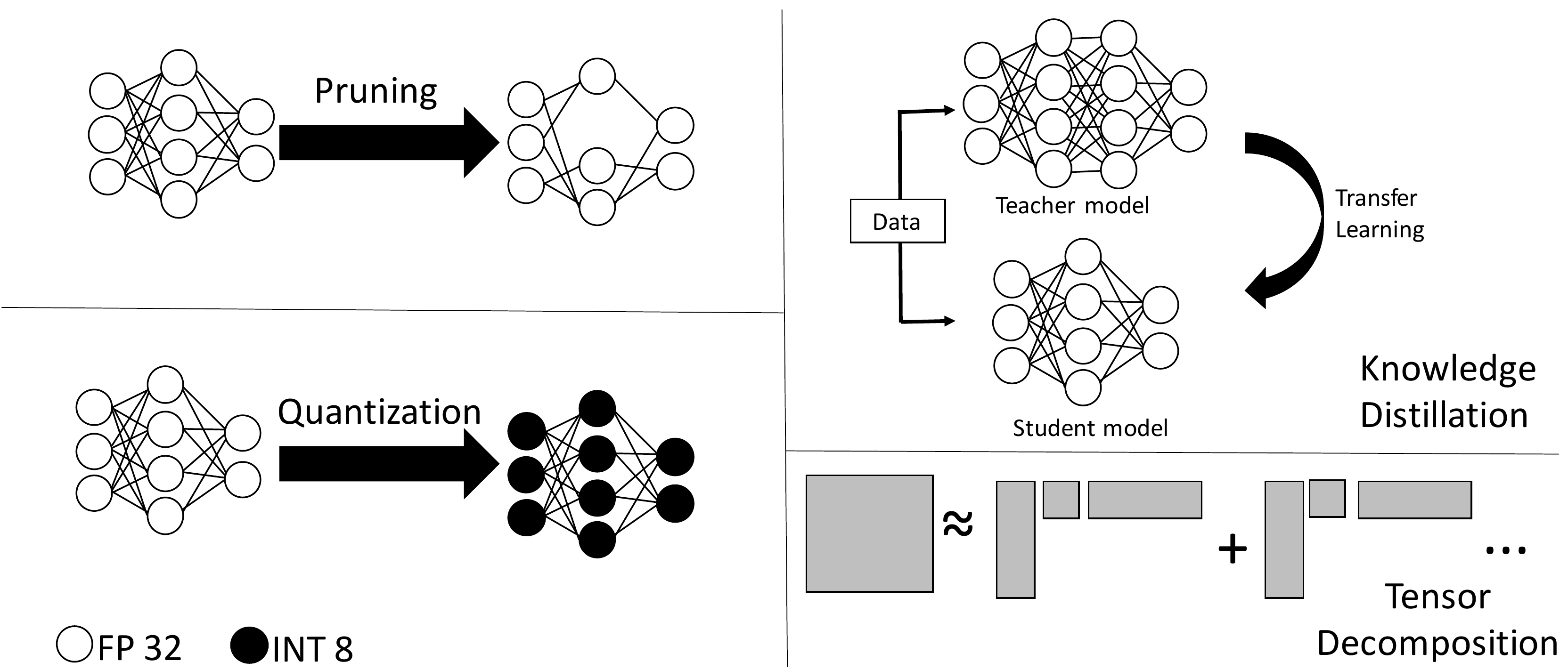}{}
    \caption{Shows an overview of the four compression techniques categories currently available in the literature. The four categories are Pruning, Quantization, Knowledge Distillation and Tensor Decomposition.}
    \label{fig:compression_methods}
\end{figure}

\section{Background and Related Work} \label{sec:background}


In this section, we cover a high-level overview of compression methods and discuss existing benchmarks.

\subsection{Compression Methods Overview}

Following Neill et al.~\cite{Neill2020} and Marino et al.~\cite{Marino2023} overviews, compression techniques beneficial for edge devices can be broken into four categories shown in Figure~\ref{fig:compression_methods} and briefly explained below:
\begin{enumerate}
    \item \textit{Pruning:} Requires the scanning and removing of neurons and connections that do not have a significant influence on the output prediction of the neural network.

    \item \textit{Quantization:} Aims at reducing the numerical representation of values in a neural network model; for example, by converting values from single-precision 32-bit floating point (FP32) numbers to 8-bit integers (INT8) or even to binarized neural networks~\cite{BNNs,TNNs}. 

    \item \textit{Knowledge Distillation:} Trains a large neural network (teacher model) on a dataset. Then a smaller neural network (student model) is trained on the same dataset whilst guided by the teacher model through transfer learning techniques to help the student model optimise achieving similar accuracy as the teacher model.

    \item \textit{Tensor Decomposition:} Involves the decomposition of large tensors by approximating them to additions and products of lower order tensors.
      
\end{enumerate}

There are other methods such as weight regularisation and conditional computing covered briefly in the literature that we neglect from the four categories described above. 
Weight regularisation involves doing modifications to the learning algorithm to allow the model to generalise better and is popularly used in retraining applications.  
We think this makes it fall better under optimisation than compression. 
However, it can also be seen as a different type of compression, where more knowledge is being compressed in to the same model without changing the architecture, structure, or the precision level of the model. 
Conditional computing is another method of compression mentioned in the literature that does not decrease size but aims at increasing runtime speed by avoiding doing certain computations for some selected input categories.  
We are interested in compression categories that can be utilised to increase efficiency on edge devices, which usually include size reduction. Therefore, weight regularisation and conditional computing are neglected in this paper.







\subsection{Compression Benchmarks}
Creating benchmarks for standardising neural network compression is currently a growing area of research. %
Microsoft Neural Network Intelligence (NNI)~\cite{nni2021} is an open-source toolkit that helps users automate the optimization of deep learning models. It supports different deep learning frameworks, mainly PyTorch and TensorFlow, allowing users to easily run experiments for optimising their models. 
One of the core parts of the Microsoft NNI toolkit is model compression. They provide a growing library of compression techniques available in the literature, such as~\cite{courbariaux2016binarized, esser2020learned, yang2022oneshot, frankle2019lottery}, making it accessible and efficient for users to experiment using SoTA compression techniques. Microsoft NNI also provides a web-based dashboard to track and visualize the results of experiments.
Neural Network Distiller (NND)~\cite{Zmora2019} is another open-source library that provides algorithmic tools from published methods to help users with neural network compression and optimisation of training post compression.
However, both NNI and NND tool-kits do not provide a comprehensive set of metrics for standardising the evaluation between compression techniques. 

Blalock et al.\cite{Blalock2020} tried to solve the lack of neural network compression standardisation in the community for pruning methods by analysing a large number of literature and generated a catalog of common issues in comparisons between pruning algorithms.
This was summarised into a set of best practices to mitigate these issues. Using these best practices they implemented ShrinkBench, a standardised library for pruning neural networks. 
Another standardised open source library is LARQ~\cite{larq}, which provides python packages for building, training and deploying binarised neural networks.   
%
For NetZIP, we focus on evaluation metrics to help engineers and researchers during testing and optimisation of different compression categories and algorithms to get a holistic view of the pros and cons. 
%
%
Whilst NetZIP compliments exiting standardised benches like LARQ, ShrinkBench, NND and Microsoft NNI with metrics for comprehensive evaluation between compression methods, compression methods can also be implemented directly in NetZIP as explained in section~\ref{sec:NetZIP_overview}. 
%
%
%

Meta research provides some relevant evaluation metrics for neural networks as part of their \textit{Slowfast}~\cite{fan2020pyslowfast} repository implementation. Whilst the metrics provided are not comprehensive for compression methods evaluation, they have provided some insight in the course of development of our paper.

An issue relevant to the context of our paper is the difficulty in generalising the removal of zeroed parameters from different pruned neural network architectures. For example, current implementations of pruning in PyTorch set a value of zero to the pruned parameters but do not automatically remove the zeros from the network structure. 
%
%
This is because most network operations are large matrix multiplications, which have been efficiently implemented in software to perform operations on fully connected graphs. Therefore, removing connections will often require custom software, unless done in a software-aware way. This problem lies outside the scope of this work.  
We direct interested readers towards DepGraph~\cite{fang2023depgraph}, which aims at generalising the approach of removing pruned parameters invariant of the architecture. 

\section{Metrics}\label{sec:metrics}
In this section, we overview different assessment metrics that serve the evaluation of DNN models and compression techniques. For each metric, we discuss how it can be computed and the information it provides for assessment. We break the evaluation metrics into five categories: Accuracy, Size, Speed, Energy, and Combined Measures.


\subsection{Accuracy}
We discuss different metrics for measuring accuracy for both object classification and detection.
We use true positives (TP) to refer to predictions that agree with ground truths; false positives (FP) for predictions which are considered false; and false negatives (FN) for ground truth annotations that were not detected.
In the context of object classifications, TPs are counted by simply matching the predicted label with the ground truth label. 
In object detection, a given bounding box is considered TP if the intersection over union (IoU), which is the ratio of overlap between a detected box and ground truth, is over a certain threshold.

\subsubsection{Top-k Accuracy}
%
Measures the proportion of samples where the ground truth class is contained in the top-k predictions of a model~\cite{petersen2022differentiable}, 
%
where $k \in \{1, ...,  K\}$ and $K$ is the max number of classes in the model being evaluated.
Two commonly used values for $k$ are 1 and 5. Top-1 accuracy in this case will only consider a prediction correct if it matches the ground truth label, whilst Top-5 accuracy will consider a prediction correct if one of the top 5 predictions of the model output matches the ground truth label. 
Top-k accuracy, for $k>1$, can be particularly useful where the model has a large number of classes, with several classes being similar to each other, or where understandable confusions exist between some classes. 
In this case, it may be challenging for the model to accurately predict the correct class, but for example, a prediction amongst the top three possibilities may still be helpful.



\subsubsection{Precision}
%
The portion of predictions that are TPs, see equation~\ref{eq:precision}. 
%
\begin{equation}
  Precision =   \frac{TP}{TP+FP}
  \label{eq:precision}
\end{equation}

\subsubsection{Recall} The portion of ground truth annotations that were predicted by the model, see equation~\ref{eq:recall}.
%
%

\begin{equation}
  Recall =   \frac{TP}{TP+FN}
  \label{eq:recall}
\end{equation}

\subsubsection{F1 Score} A metric that combines both precision and recall into a single metric. The F1 score is the harmonic mean of precision and recall, given by equation~\ref{F1_score}.
The F1 score provides a way to balance the trade-off between precision and recall, as both measures are important for different aspects of model performance. A high F1 score indicates that a model has both high precision and high recall.

\begin{equation}
  \text{F1 Score} =   \frac{2 \cdot \text{Precision} \cdot \text{Recall}}{\text{Precision} + \text{Recall}}
  \label{F1_score}
\end{equation}

\subsubsection{Mean average precision (mAP)} A metric that takes into account both precision and recall of a model. 
The metric is calculated by summing the average precision ($AP$) of each class $k$ over the number of classes $K$, as shown by equation~\ref{eq:mAP}.  
%
The AP of each class is calculated by plotting precision for different recall values and calculating the area under the curve (AUC)\footnote{To plot the precision-recall graph, an IoU threshold needs to be selected. In the case where an optimal IoU threshold is not selected or unknown, then the AUC can be calculated for several IoU thresholds and averaged to give an AP representative of a range of IoU thresholds.}.
Compared to F1 score, mAP is more sensitive to imbalances in the number of training samples between different classes, whilst the F1 score is insensitive to these imbalances in training samples. On the other hand mAP is more expensive to compute compared to the F1 score.
%

\begin{equation}
    mAP = \frac{1}{K} \sum_{k=1}^{k=K} AP_k
    \label{eq:mAP}
\end{equation}

\subsection{Speed}
\subsubsection{Inference Latency} The time taken for the model to output its prediction. 
Whilst inference latency may give the most practical real-time value for operational speed estimates and machine-specific comparisons, it is affected by resource utilisation at the time of measurement. 
This makes the measured inference latency variant to the level of resource utilisation, and thus can be less informative about the benefit of compression. This is especially the case if experiments are run on different machines or at different times where the level of resource utilisation may vary. 
This is analogous to an issue in simulation-based testing, where variances in results for the same test can vary significantly due to resource utilisation changes~\cite{determinisim}.

\subsubsection{Number of Operations (OPs)}
The number of operations (OPs) required to process a given input can provide a hardware-agnostic measure of the improvement in speed gained by a compression method.
It does not provide any information for quantization as this family of compression does not reduce the number of OPs, just the amount of hardware needed to carry out a given operation~\cite{Budgett2022}. 
%
%
Floating point operations (FLOPs) are often referred to in literature instead of OPs due to most neural network processing being done with floating point operations. However, to generalise between floating point and quantised operations, we will use OPs.
Furthermore, most operations in a neural network are multiply-accumulator operations (MACs), which perform one multiplication and one addition. 
Given that MACs refer to the majority of neural network operations and some hardware are tuned to processing MACs, this measure is often used instead of OPs.

\subsubsection{Compression and Hardware Agnostic Theoretical Speed (CHATS)}
\begin{figure}[H]
    \centering
    \begin{subfigure}{0.3\textwidth}
        \includegraphics[width=1\textwidth]{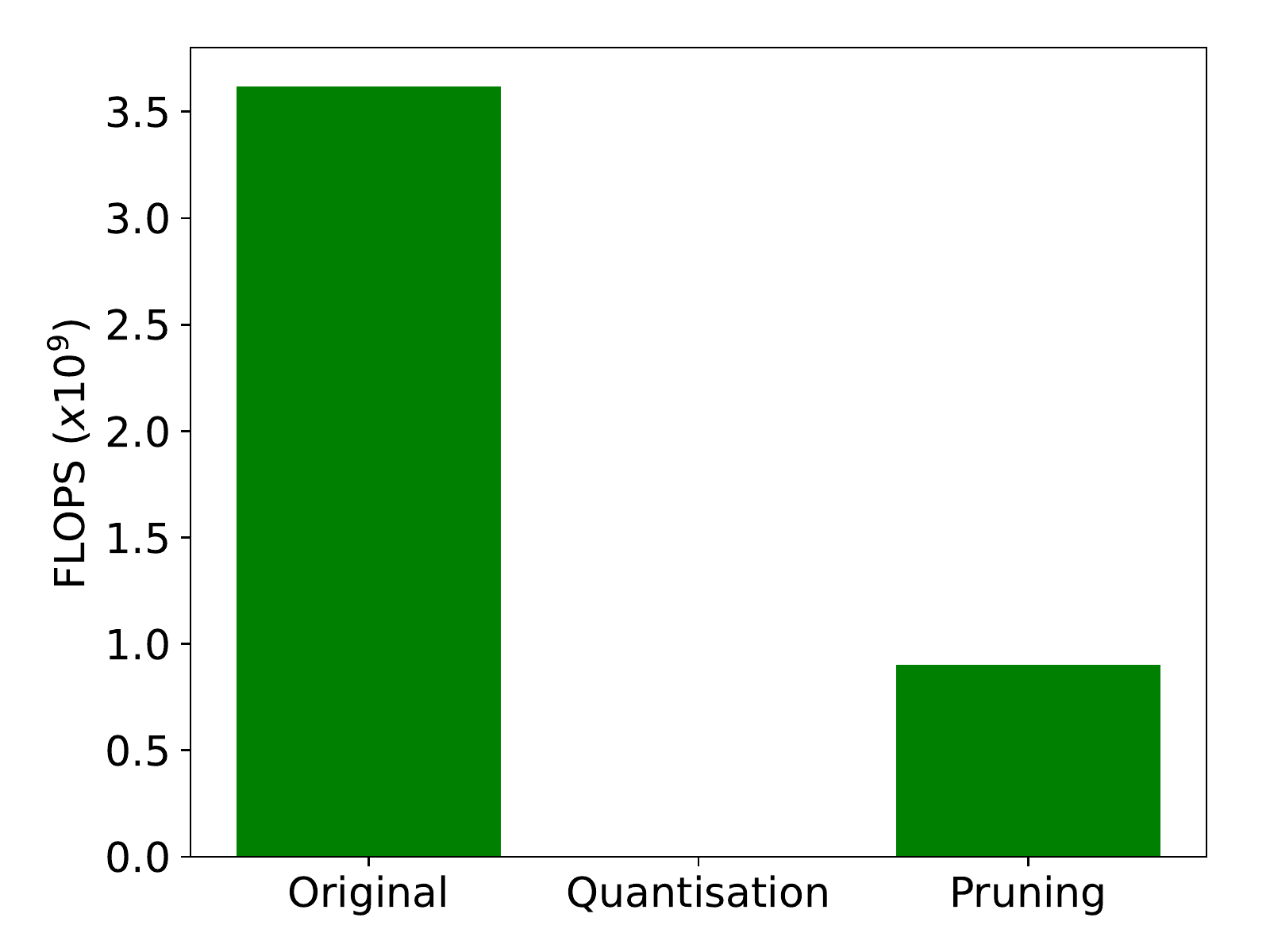}
        \caption{FLOPs}
    \end{subfigure}
    \begin{subfigure}{0.3\textwidth}
        \includegraphics[width=1\textwidth]{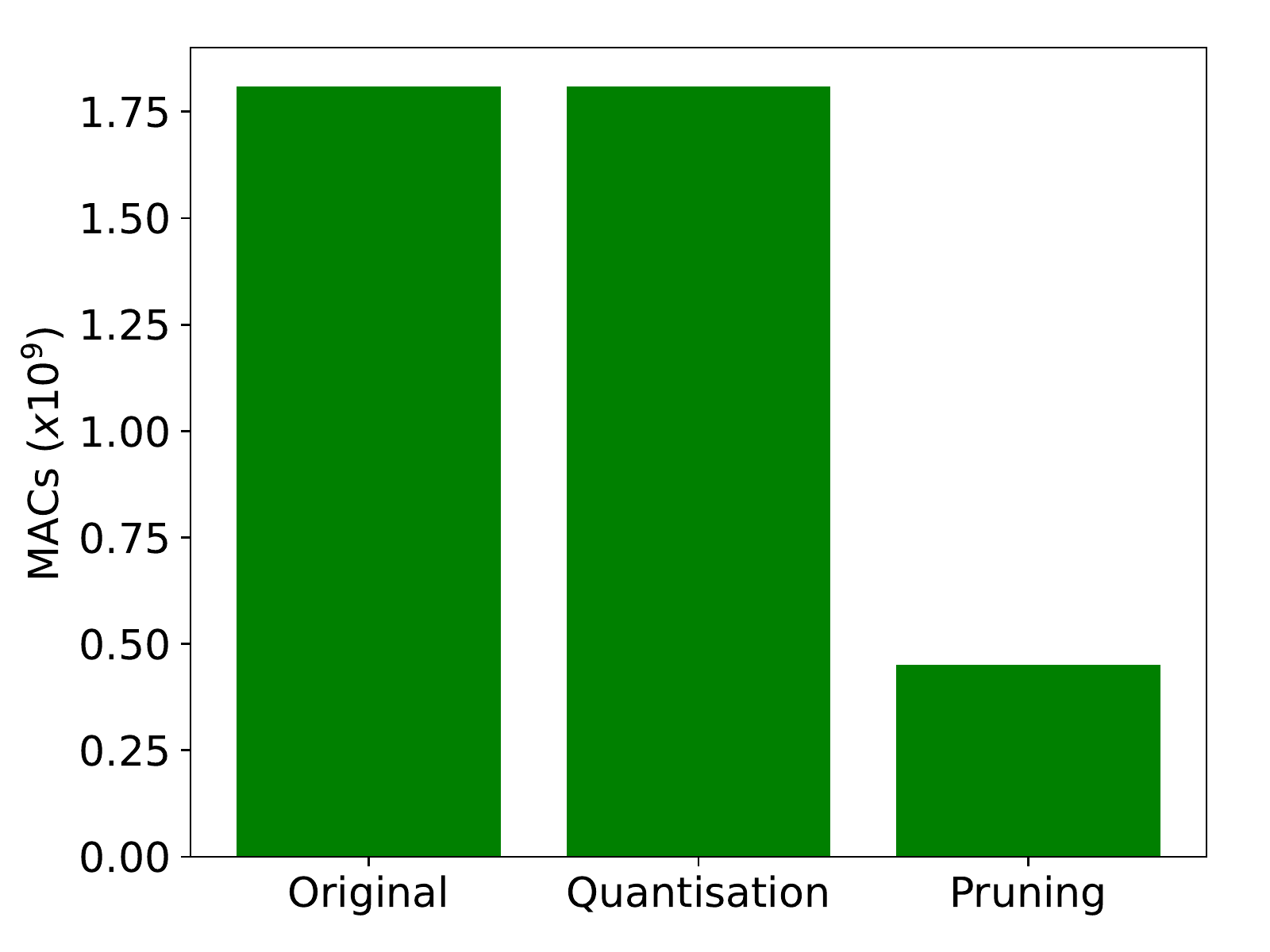}
        \caption{MACs}
    \end{subfigure}
    \begin{subfigure}{0.3\textwidth}
        \includegraphics[width=1\textwidth]{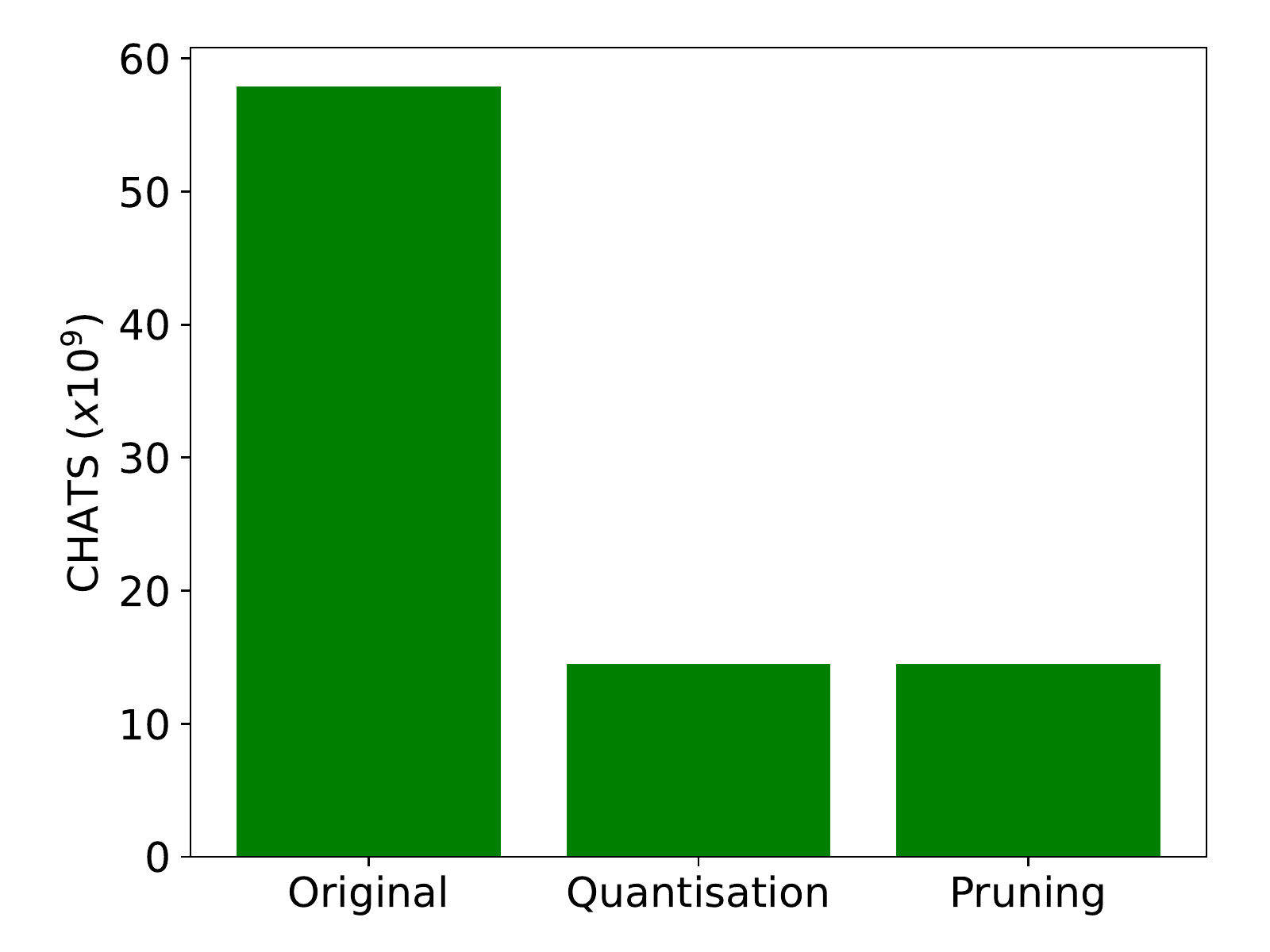}
        \caption{CHATS}
    \end{subfigure}
   
    \caption{Plots showing the measurement of theoretical speed using FLOPs, MACs and CHATS for pruning and quantitation of Resnet18. Quantization is done from FP32 to INT8 and pruning is done at sparsity level of 75\%. For both quantization and pruning the drop in theoretical speed compared to the original mode should be about 1/4$^{\text{th}}$. As it can be seen FLOPs and MACs fail to report this theoretical speed reduction correctly for  quantization but CHATS does this successfully for both quantization and pruning. }
    \label{fig:comparing-CHATS}
\end{figure}

Given the current limitations of previously discussed speed metrics, relating to poor generalisation between compression techniques and sensitivity to resource utilisation variances, we introduce the \textit{CHATS} metric.
CHATS aims at providing a theoretical speed quantity that is agnostic to the compression technique, and simultaneously not affected by the resource utilisation levels of hardware. See Figure~\ref{fig:comparing-CHATS} for comparison between FLOPs, MACs and CHATS.
The metric succeeds in providing such theoretical speed by combining OPs with the bit width representation used by a model, see equation~\ref{eq:CHATS}.

The number of binary digits to be processed scales as the square of the bit width for multiplication, whilst for addition it scales linearly with the bit width. 
In DNNs, where both multiplication and addition may exist, the actual reduction in processing binary digits as a result of reducing the bit width will scale somewhere between the square of the bit width and linearly with the bit width. 
Leading GPU developers, such as NVIDIA, often report improvements in theoretical speeds as a linear relationship with bit width, e.g.~\cite{NVIDIA_TURING_GPU_ARCHITECTURE_2018}.  
Similarly, we also adopt a linear relationship with bit width when calculating CHATS. We introduce a hardware specific constant, $\zeta$, when calculating the speedup ratio for CHATS (see section~\ref{combined_measures}) to act as a scaling coefficient dependent on the hardware. 
For example, on the results of NVIDIA Turing Tensor Cores hardware shown in~\cite{NVIDIA_TURING_GPU_ARCHITECTURE_2018}, the speedup ratio scales linearly with the bit width and has a hardware constant $\zeta = 4$. 

Furthermore, as technology of DNNs tailored hardware advances to prioritise efficiency, where multiplication operations dominate, e.g.~\cite{Budgett2022}, we may find that CHATS scales as the square of the bit width. Investigating this further remains as future work.

 
%

\begin{equation}
    CHATS = OPs \cdot bit width
    \label{eq:CHATS}
\end{equation}

\subsection{Size}
\subsubsection{Disk size}
The space required to store the model on a computer can be of significance especially when models are large, e.g. ChatGPT-3~\cite{Brown2020}. Compression techniques can help in significantly reducing the model size so less disk space is required. For example, quantization of model parameters from FP32 to INT8 numerical representation decreases disk size required to save the model typically by a quarter. 
%

\subsubsection{Parameters Count}
Another way of measuring the size of a model is by counting the number of parameters in the compressed model compared to the uncompressed version. This method can be useful when comparing compression techniques where parameters are actually removed from the model, e.g. pruning. 
The usefulness of parameters count will fall short when it comes to quantization techniques, where the number of parameters stay the same but the numerical representation changes.
In these cases, using the disk size as a metric to compare between compression techniques can be a more informative measure than parameters count.
Nevertheless, parameters count can be of great use during early stage assessment of pruning techniques.
In many of the popular pruning libraries implementations, e.g. PyTorch~\cite{paszke2017automatic}, typically the pruned parameters are zeroed but not removed from the model.
The proportion of the uncompressed model parameters that are zeros after compression is known as \textit{sparsity}.
%
Sparsity, can serve to show how much reduction in disk size will be achieved using different pruning techniques, prior to committing to the manual process of removing pruned parameters. 
%


\subsubsection{CPU/GPU utilisation}

\begin{figure}[h]
    \centering
    \includegraphics[width=0.8\textwidth]{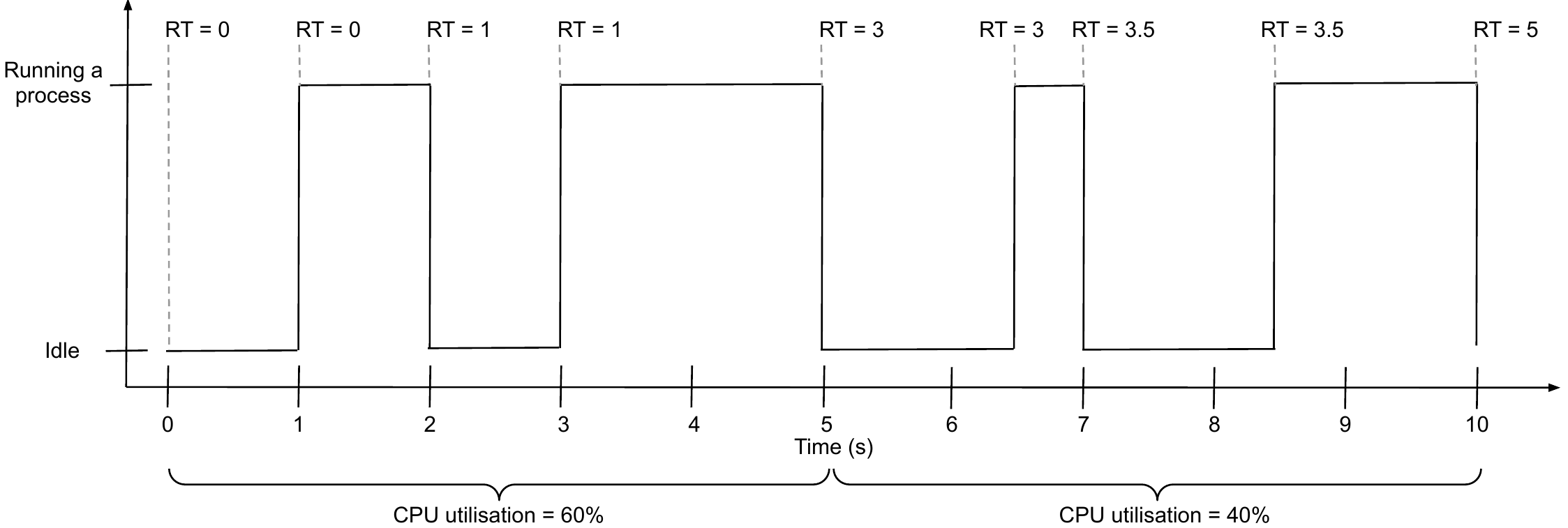}
    \caption{Illustration of CPU switching between running a process and being idle.}
    \label{fig:cpu_utilisation_illustration}
\end{figure}

When a CPU is not running a program, it is idle. The running total (RT) is the total time the CPU spent on running programs, i.e. not being idle, through-out the active time of the computer. 
The CPU utilisation is the percentage of time the CPU is not idle, calculated by subtracting two samples of RT and dividing by the time between taking the two samples.
Figure~\ref{fig:cpu_utilisation_illustration}, shows an illustration for CPU switching between running a program and being idle. If sample measurements are taken every five seconds, then the CPU utilisation for the initial 5s was 60\% then dropped to 40\%. 
GPU utilisation will take  a similar approach of measuring time spent on processing requests against idle time to calculate utilisation. 
In terms of how CPU and GPU utilisation can be measured practically, \texttt{psutil} \cite{psutil2019} and \texttt{nvidia-smi}~\cite{NVIDIA_SMI} are popular libraries that provide these measurements, respectively.

Measuring the reduction in computational resources used by the model to output a prediction is an important factor to assess when investigating the trade-offs between different compression techniques.
When taking measurements for utilisation, other programs should be terminated and only the model outputting predictions should be the running process. In most computers, however, there are many background processes that need to run for the computer to operate. 
In which case, the best estimate is to subtract the utilisation measurement before and during making predictions.

\subsubsection{RAM usage}
RAM (Random Access Memory) usage in inference is a measure of paramount importance to report. 
Knowing how much reduction in RAM usage is achieved by different compression techniques can be advantageous in determining specifications of hardware needed for a model to operate on resource-constrained devices.
Libraries such as \texttt{psutil} \cite{psutil2019} can be used to determined RAM usage.
Similar to CPU utilisation, when taking measurements of RAM usage, other programs should be terminated other than the inference model. 
%






\subsection{Energy consumption}
Energy consumption can be calculated by plotting power against time and the area under the plotted curve is the energy consumed.
In other words, energy consumed $E$ is the integration of power $P(t)$ as function of time $t$ over the duration $t_0$ to $T$, given by Equation~\ref{eq:energy}.

\begin{equation}
    E = \int_{t_0}^{T} P(t) \,dt = \sum_{t=t_0}^{t=T} P(t)\cdot  dt
    \label{eq:energy}
\end{equation}

\begin{figure}[h]
    \centering
    \includegraphics[width=0.6\textwidth]{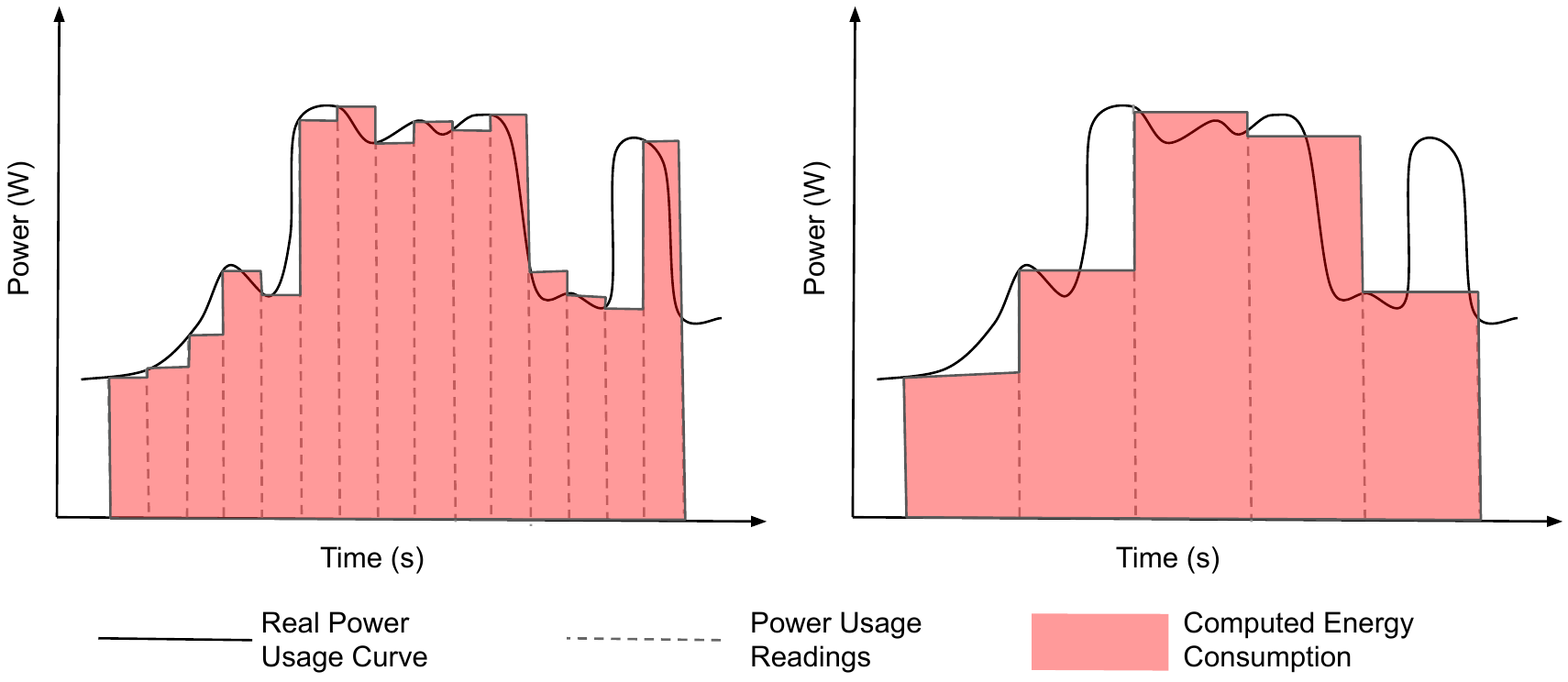}
    \caption{Illustration of energy measurement from fine (left plot) and coarse (right plot) power reading intervals.}
    \label{fig:energy_illustration}
\end{figure}

Figure~\ref{fig:energy_illustration} illustrates the measurement of energy consumption using fine and coarse sampling rates of power. It is important that sampled power measurements are fine enough to sufficiently capture detail in the energy consumption profile. For example, as illustrated by the right diagram in Figure~\ref{fig:energy_illustration}, a spike in the energy consumption profile was missed due to the coarse power reading intervals.

Measuring the energy consumption of a neural network depends on various factors such as the hardware platform and the workload being executed. 
For reliable comparisons of energy consumption, one needs to ensure that the same hardware is used with the same resource utilisation setting.
One common python library used to measure energy consumption in computers employing Intel processors is pyRAPL, which is a Python toolkit used to measure the energy footprint of the CPU~\cite{pyRAPL_repo}.  
Whilst pyRAPL may give good indicative estimates of energy consumption, its measurements are reliant on estimation models instead of live measurements of power. 
This may be useful for trying to improve the energy utilisation of a machine but may have limitations when budgeting for energy requirements, which is critical in some application, e.g. space applications.
Using pyRAPL also limits measurement to hardware only employing Intel processors. 

A more generic  way is to use power measurement tools that can connect between the power source and the hardware platform to measure the power consumption in real time. 
This way of measuring energy is more informative for applications requiring energy budgeting. 

\subsection{Combined Measures} \label{combined_measures}
We use this section to capture derived metrics from the aforementioned metrics. 
%
Success of compression techniques is often reported in the literature in the form of improvement ratios~\cite{He2018,Blalock2018,Menghani2023}. Two popular improvement ratios are \textit{compression ratio} and \textit{(theoretical) speedup ratio}.
%
The speedup ratio refers to the reduction in computation time that can be achieved by using a compressed model, relative to the original uncompressed model. Typically calculated as a ratio between the computation time of the uncompressed model and the compressed model (see Equation~\ref{speedup_ratio}). When this speedup ratio computation is carried out using a non-temporal metric, e.g.(OPs, FLOPs, MAC), then it is referred to as \textit{theoretical speedup}. 
We incorporate the hardware specific constant, $\zeta$, in the speedup ratio to provide means of correcting the speedup ratio based on the operational hardware. This hardware specific constant is determined empirically. 

%
\textit{Compression ratio} is another derived metric based on size reduction of the model. It refers to the reduction in model size achieved by a compression technique relative to the size of the original uncompressed model (see Equation~\ref{compression_ratio}). 
We extend this further to include the \textit{efficiency ratio} (Equation~\ref{effeciency_ratio}) and \textit{performance ratio} (Equation~\ref{performance_ratio}), which aim at reporting improvements in  energy consumption and accuracy, respectively, between the compressed and uncompressed models.

\begin{equation}
  \text{Speedup ratio} =   \zeta \cdot\frac{\text{original speed}}{\text{compressed speed}}
  \label{speedup_ratio}
\end{equation}

\begin{equation}
  \text{Compression Ratio} =   \frac{\text{original size}}{\text{compressed size}}
  \label{compression_ratio}
\end{equation}

\begin{equation}
  \text{Efficiency Ratio} =   \frac{\text{original energy consumption}}{\text{compressed energy consumption}}
  \label{effeciency_ratio}
\end{equation}

\begin{equation}
  \text{Performance Ratio} =   \frac{\text{compressed accuracy}}{\text{original accuracy}}
  \label{performance_ratio}
\end{equation}

Reporting improvement ratios alone is not enough, as the ratios can be calculated using different metrics; for example one may calculate the compression ratio using disk size, whilst another can use parameters count, or even CPU/GPU usage. Similarly, one may use inference latency, number of OPs, or MACs to compute speedup ratio.
Therefore, when reporting improvement ratios, one needs to outline what metrics were used in their computation to allow for reliable comparisons between different works.

\subsubsection{Overall Compression Success (OCS)}
We introduce the Overall Compression Success (OCS) metric. The metric combines the different improvement ratios into one metric summarising the overall success of a compression technique. 
OCS is shown by equation~\ref{OCS}, where P, S, C and E are the calculated performance, speed, compression and efficiency ratios, respectively. 
%
%
The OCS metric is set in a way to output a positive value if there is an overall improvement, a negative value if there is an overall  worsening. OCS will output a value close to zero if there is no change from the uncompressed model, the trade-offs cancel out, or the compressed model accuracy has dropped significantly. 

OCS can be a useful initial filter of low performing compression methods. 
As the number of combinations of neural networks and compression methods scale up, comparisons using several metrics can be inefficient and challenging especially when there are trade-offs to be made. 
The OCS metric gives a single score of the holistic advantage of each compression technique. This is particularly beneficial in commercial setups. 
A prime example of such usage in commercial environments can be seen in how NVIDIA compares between its numerous hardware platforms via using a bar chart plotted using a single scoring approach~\cite{NVIDIA_TURING_GPU_ARCHITECTURE_2018}. 
The filtered methods can be then further studied using radar plots displaying the different efficiency metrics, as we suggest and shown in section~\ref{sec:results}.


\begin{equation}
  \text{OCS} =  P^2 \cdot \Bigl((P-1) + (S-1) + (C-1) + (E-1) \Bigl) 
  \label{OCS}
\end{equation}

\section{NetZIP Overview} \label{sec:NetZIP_overview}

\begin{figure}[]
    \centering
    \includegraphics[width=0.7\textwidth]{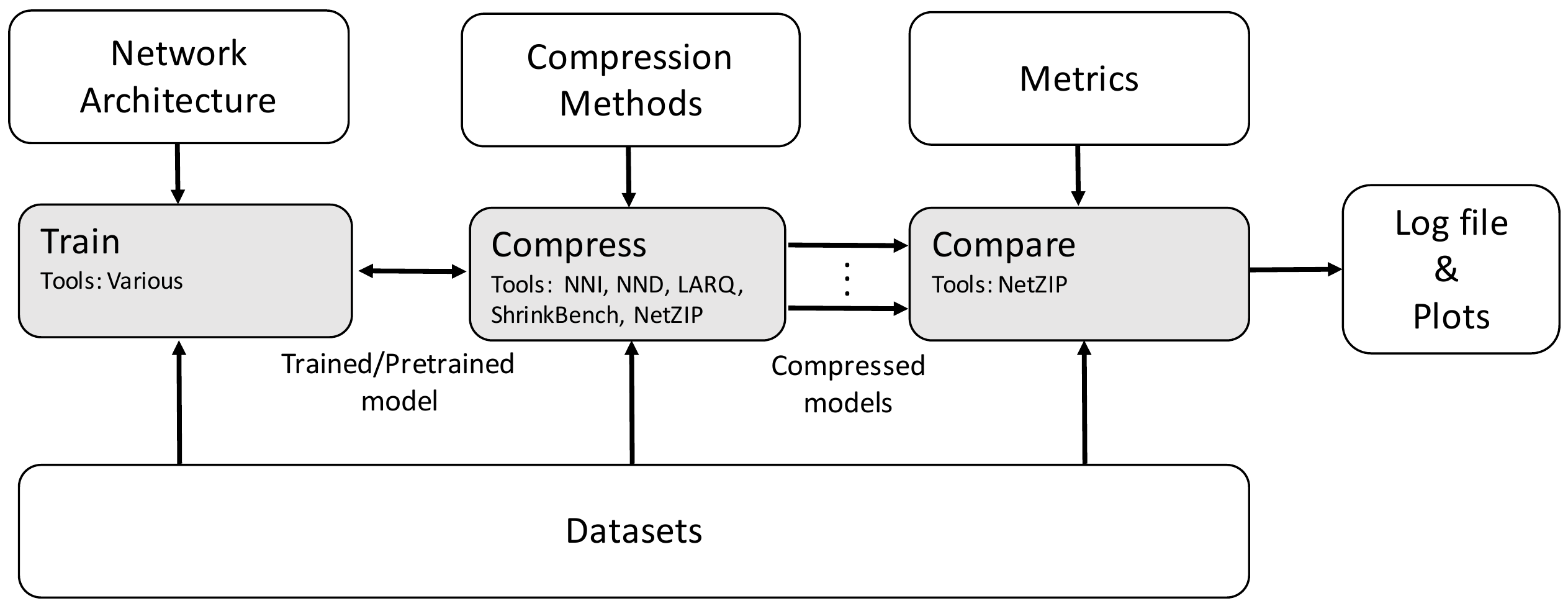}
    \caption{NetZIP overview}
    \label{fig:Netzip}
\end{figure}

We introduce NetZIP, a library that provides a suite of metrics for evaluating the gains and losses in performance between different compression techniques.
The suite of metrics is based on the five evaluation categories reviewed in  section~\ref{sec:metrics}.
The goal of NetZIP is to provide a unified implementation of assessment metrics to ease and standardise the evaluation of compression techniques. This allows for more reliable and unified comparisons across works by different researchers.
Figure~\ref{fig:Netzip} shows an overview of the three major stages involved in NetZIP: Train, Compress, Compare. These three stages represent the generic approach a neural network needs to go through to find the most optimised compressed model. 

Training, entails optimising the parameters of a chosen neural network architecture to maximise accuracy of predictions on a selected dataset. There are many tools and libraries in the literature that are used collectively to achieve highly optimised models.
Compressing involves using different compression methods to increase the efficiency and reduce overparameterization of the model. Note that often one may need to carry further training after compression to optimise the model further. We have reviewed several powerful libraries that implements different methods for compression in section~\ref{sec:background}.
The next stage is comparing between the different compressed models. The availability for an evaluation platform agnostic of the used compression technique lacks in the literature. 

The main contributions of NetZIP are the metrics and evaluation between different compression techniques that it provides regardless of the type of compression. 
In contrast with other libraries, for example, Shrinkbench and LARQ also provide aspects for evaluation but only specific to one category of compression like pruning or binarisation only.
In addition, NetZIP also allows users to implement their own training and compression methods or integrate it with other more developed libraries like LARQ, NNI, NND, and Shrinkbench; making NetZIP employable in and independent or complimentary fashion. 
%
%
With community support, our goal for NetZIP is to improve and grow overtime, to include more metrics and baseline experiments as research on neural network compression evolves. 
%
%

%
%
%
%


\subsection{Visualisations and analysis}
\begin{figure}[]
    \centering
    \includegraphics[width=0.75\textwidth]{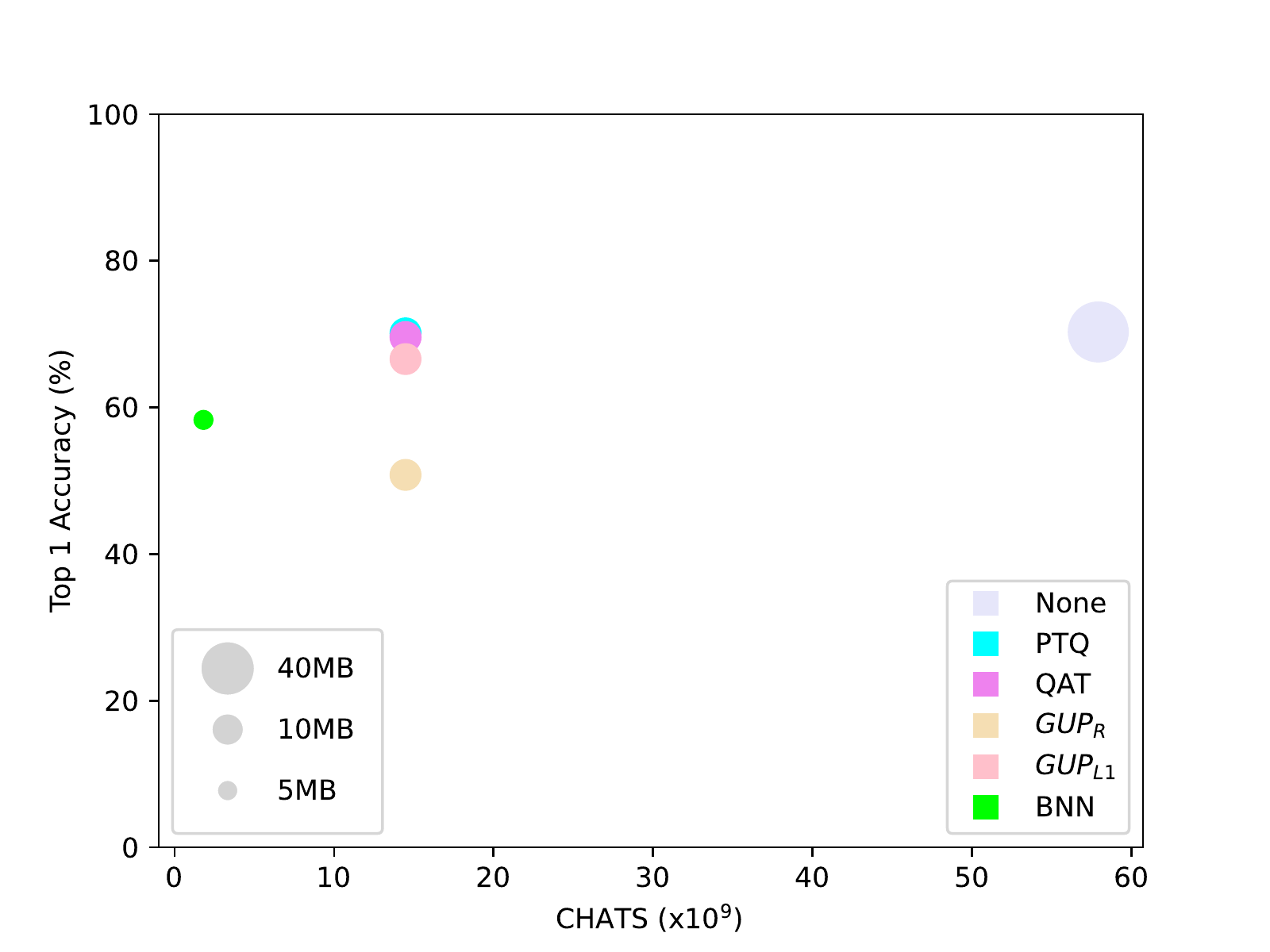}
    \caption{Popular plot format for comparing between different compression techniques, implemented for case study 1. The x-axis shows speed and y-axis showing accuracy. The marker size indicates the size of the model and the colour shows the different compression techniques.}
    \label{fig:popular_comparison_plots}
\end{figure}

%
A popular way of visualising the benefit in performance of different compression techniques is by plotting the results in Figure~\ref{fig:popular_comparison_plots}'s format; also see e.g.\cite{Bannink2020}.
This plot provides a comprehensive way of visualising the benefits of different compression techniques when comparing accuracy, speed, and size. 
However, if  more factors are to be enrolled in the comparison, such as energy efficiency, it becomes difficult to utilise Figure~\ref{fig:popular_comparison_plots}'s format.
Radar plots on other hand, allows for more axes to be enrolled whilst still providing a comprehensive view of the benefits of the different compression techniques.
As part of NetZIP, we output log files and radar plots to visualise comparisons between different compression techniques.
If comparing between a large number of compression techniques, radar plots may become very crowded. In this case we propose the utilisation of the OCS metric to filter out the compression methods yielding the highest output. These filtered techniques can then be visualised using radar plots. 
%

\section{Case Studies} \label{sec:casestudies}
We use three case studies implemented using NetZIP to showcase some of the evaluation metrics reviewed and the novel metrics we contributed in this paper. 
Table~\ref{tab:casestudy_summary}, shows a summary of the case studies, including the neural network architectures, datasets, compression techniques, and the hardware used in our case studies.
Table~\ref{tab:Hardware_summary} shows the hardware specifications. 
In all of our evaluations we eliminated the usage of GPU and used only CPU, to maintain fair availability of resources in our experiments, especially since some compression implementations currently only operate on CPU. 
The majority of metrics used in the assessment are the same for all case studies, as can be seen from results in tables~\ref{tab:resnet-imagnet1k}, ~\ref{tab:Yolov5s-COCO}, and \ref{tab:various-BNN-Imagnet1k}. The only different assessment metric is the accuracy metric for case study 2; we used mean average precision instead of top-1 accuracy.

\subsection{Case Study 1}
In the first case study, we use a classic object classification as our theme, using ResNet18~\cite{he2015deep} architecture with ImageNet1k~\cite{imagenet_cvpr09} dataset. 
This case study was run on a powerful PC (see table~\ref{tab:Hardware_summary}).
We used five compression techniques available in PyTorch and LARQ~\cite{larq}. Two compression techniques use quantization, two use pruning, and one uses binarization: 
\begin{enumerate}
    \item Post Training Quantization (PTQ), where the model is trained on the dataset and then quantised.
    
    \item Quantization Aware Training (QAT), where the model is trained on the dataset, quantised and then further training is done to try and recover any drops in accuracy resulting from quantization.

    \item Global Unstructured Random Pruning (GUP$_R$), where parameters through out the model are pruned randomly until the required level of sparsity is achieved. 

    \item Global Unstructured L1-norm Pruning (GUP$_{L1}$), prunes the least significant parameters in the model globally until the required level of sparsity is achieved. 

    \item Binary Neural Network (BNN), replaces floating-points with binary. In this case we have used LARQ~\cite{larq} to get a binarized Resnet18.    
\end{enumerate}

\subsection{Case Study 2}
The second case study focuses on low energy devices. We use object detection using YOLOv5s~\cite{Jocher_YOLOv5_by_Ultralytics_2020} to output predictions for the COCO~\cite{lin2015microsoft} dataset.
We use the compression methods already integrated within the YOLOv5 repository~\cite{Jocher_YOLOv5_by_Ultralytics_2020}, which are limited to post-training quantization using TensorFlow lite (PTQ-TFlite) to FP16 and INT8 numerical representations. We also included in our comparison the Tensorflow (TF) FP32 version of the YOLOv5s architecture. 
In this case study, we have run experiments using Raspberry Pi 4 Model B, as a means of studying the benefits of compression on hardware with limited resources.

\subsection{Case Study 3}
The third case study aims at showcasing the Overall Compression Success (OCS) metric. 
In this case study we use different quantised and binarized neural networks to find which is best for classifying Imagenet-1k tasks. Our baseline to compare the success of compression is the uncompressed Resnet18 model from case study 1. See table~\ref{tab:various-BNN-Imagnet1k} for a list of the used compressed neural network models. The compressed models are either quantised models using NetZIP or binarized models based from LARQ~\cite{larq} framework.



\begin{table}[]
    \centering
    \begin{tabular}{| c | c  | c | c | c | c |} 
        \hline
        Case Study No. & Theme & Network Architecture & Datasets & Compression Techniques & Hardware  \\
        \hline        
        1 & Object Classification  & ResNet18 & ImageNet1k & PTQ, QAT, GUP$_R$, GUP$_{L1}$, BNN & PC \\ 
        2 & Object Detection  & YOLOv5s & COCO  & PTQ-TFlite  & RasPi 4 \\
        3 & Object Classification  & Various (see table~\ref{tab:various-BNN-Imagnet1k}) & ImageNet1k & BNN & PC \\ 
        \hline
    \end{tabular}
    \caption{Case Studies Summary}
    \label{tab:casestudy_summary}
\end{table}
\begin{table}[]
    \centering
    \begin{tabular}{| l  | l | l | l |} 
        \hline
        Name & Product Name & Specifications & Operating System \\
        \hline        
        PC  & Dell Alienware Desktop & 64GB RAM
        & Linux Ubuntu 18.04.4 LTS (64-bit)\\
        RasPi 4  & Raspberry Pi 4 Model B & 4GB RAM &  Linux Ubuntu 22.04.2 LTS (64-bit) \\
        \hline
    \end{tabular}
    \caption{Hardware Specifications Summary.}
    \label{tab:Hardware_summary}
\end{table}

\section{Results and Discussion} \label{sec:results}
Tables~\ref{tab:resnet-imagnet1k}, ~\ref{tab:Yolov5s-COCO}, and ~\ref{tab:various-BNN-Imagnet1k} show a summary of the results for case studies 1, 2 and 3, respectively. 
Figure~\ref{fig:Radar-spider} shows radar plots summarising the results for case studies 1 and 2. Figure~\ref{fig:various} summarises results for case study 3.

\subsection{Assessment of accuracy and speed}
For case study 1, BNN gave the best overall compression success. 
However, PTQ provided the least drop in accuracy of 0.2\% compared to the uncompressed model,  whilst GUP$_R$ had the most significant drop in accuracy of about 19.5\%. 
%
In terms of processing speed, assessed using latency, compression using quantization improved the inference latency by $\times 2$. However, no changes in speed are seen for pruning techniques, and for BNN the latency worsened significantly by a factor of $\times 5$. 
Note that inference latency is sensitive to hardware resource utilisation and the library implementation. For example, the BNN model is based on LARQ which uses TensorFlow, whilst the other compression approaches are based on PyTorch. 
Therefore, we resort to observing theoretical speeds provided by MACs and CHATS metrics, which are insensitive to resource utilisation variations and library implementations. 
MACs suggests there should be no change in speed for any of the compression techniques, but using the CHATS metric we introduced, it can be seen that it suggests that inference time should decrease for quantization techniques but should stay the same for pruning techniques.
As discussed previously in our related works section, current implementations of pruning are limited to setting pruned parameters to zero but they are not actually removed from the model. Therefore, we see changes in accuracy due to pruning, but not in speed, size, or efficiency metrics.
However, we know that pruning was done with a sparsity level of 75\%, so for the speed and size theoretical metrics they should see an improvement of about $\times 4$, similar to quantization. We have include these assumed theoretical improvements in brackets for the pruning techniques in Table~\ref{tab:resnet-imagnet1k}.
%

In case study 2, there were no significant drops in accuracy, the highest drop of 1.9\% was incurred by PTQ TFlite (INT8) compression technique.   
%
For the second case study, the latency and CHATS decreased as expected, with increasing quantization levels. 
However, interestingly the TensorFlow FP32 model had the least inference latency. 
We speculate that this may be a result of different library implementations compared to PyTorch. 
Investigating these speculations are beyond the scope of this paper, however it is important that these factors are investigated further in real life applications.
%

\subsection{Size compression assessment}
In case studies 1 and 2, compression using quantization or binarization decreased the disk size of the model but has not changed the parameters count as expected. 
For pruning in case study 1 this decrease can not be noticed, but to showcase the potential provided by pruning in size reduction we have included the parameters count for non zero parameters between brackets.
%
Reduced hardware usage by compression was only observed for CPU utilisation and gigabytes of RAM usage for binarization in case study 1, but no change for other quantization and pruning methods.
On the other hand, on less powerful hardware used in case study 2, significant reduction in hardware usage can be observed for the compressed (quantised) models.

\subsection{Energy efficiency assessment}
Compression caused energy consumption improvements in both case studies 1 and 2. 
In case study 1, quantization provided a significant drop in the energy consumption. 
The binarized model on the other hand used the least power but spent the most energy overall, this is due to the latency in outputting predictions was higher.
In case study 2, even though the TF (FP32) implementation had better speed compared to TFlite (INT8), the energy efficiency of the PTQ TFlite (INT8) was best.  
This observation is interesting as one may expect the INT8 implementation to be both quicker and more energy efficient. 
Field et al.~\cite{Field2014} reported a similar observation, where shorter bits save more energy but take more execution time than longer bit representations. 
This may perhaps be resolved through reconfiguration of hardware, however, it is beyond the scope of this paper to investigate technically how this can be achieved.

\subsection{Benefit of Overall Compression Success (OCS) metric}
\begin{figure}[]
    \centering
    \begin{subfigure}{0.49\textwidth}
        \includegraphics[width=1\textwidth]{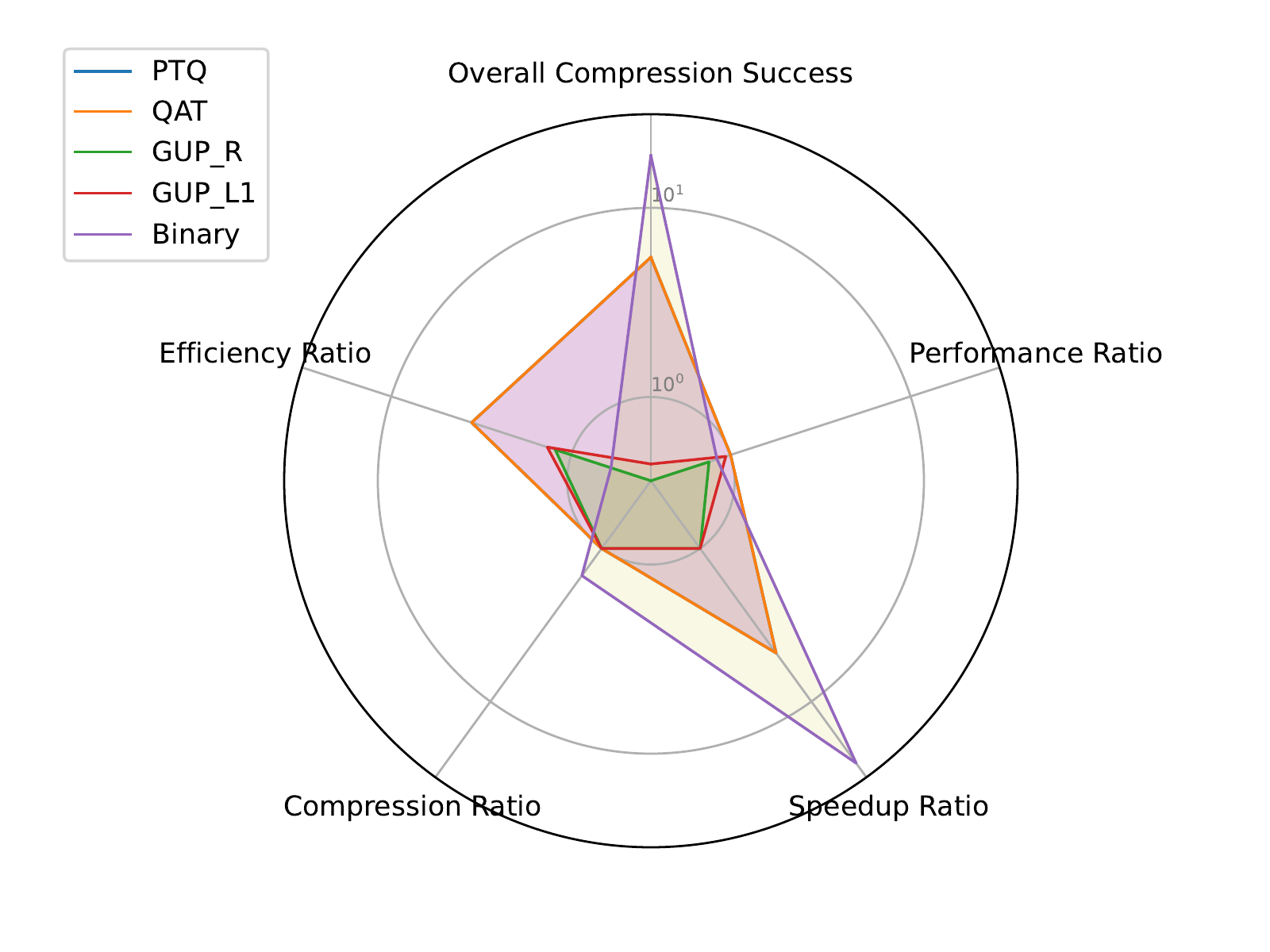}
        \caption{}
        \label{fig:spider-case-study1}
    \end{subfigure}
    \begin{subfigure}{0.49\textwidth}
        \includegraphics[width=1\textwidth]{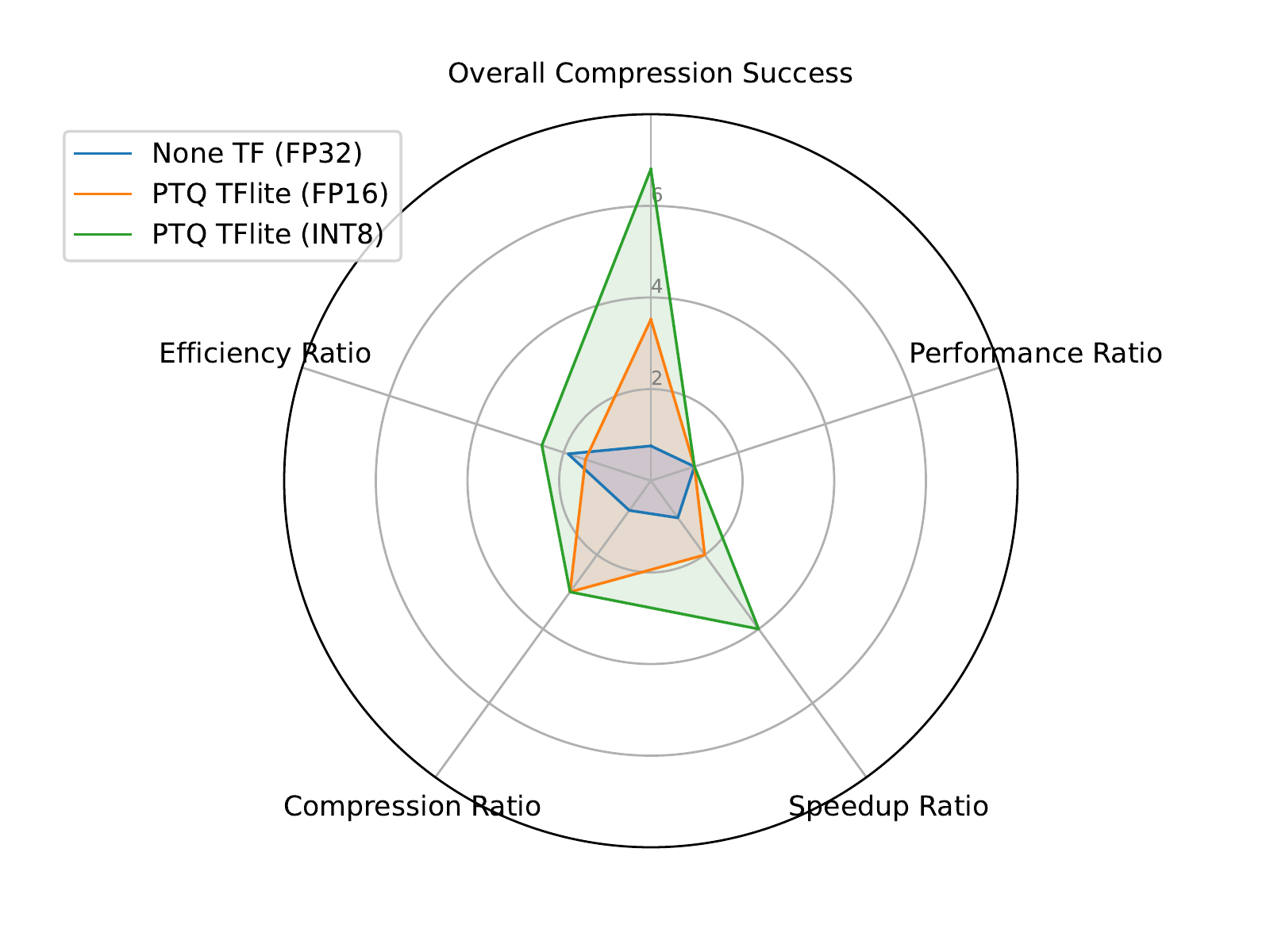}
        \caption{}
        \label{fig:spider-case-study2}
    \end{subfigure}
   
    \caption{Radar plots summarising improvement ratios for (a) case study 1 and (b) for case study 2. The following summarises the metrics used in calculating the improvement ratios: Performance Ratio (Top1 for object classification and mAP for object detection),  Speedup Ratio (CHATS), Compression Ratio (CPU utilisation), Efficiency Ratio (Energy).}
    \label{fig:Radar-spider}
\end{figure}

As can be seen, comparing benefits of compression is a multivariate problem, which can be difficult to quickly digest when comparing different techniques. 
The Overall Compression Success (OCS) metric can assist by giving an idea of which techniques perform best overall. This is more informative when viewed a long with the other improvement ratios. 
Therefore, we choose to view the overall compression success using radar plots, which provides an optimal way for observing overall compression success and the trade-offs made (see figure~\ref{fig:Radar-spider}) in the different assessment categories. 
The OCS and each improvement ratio are represented by a separate axis in figure~\ref{fig:Radar-spider}. Data points for each improvement ratio are plotted on the axes and connected to form a polygon. 
The value of OCS and the shape of this polygon represent a more tangible approach for assessing the best compression technique. 

Using the radar plots in figure~\ref{fig:Radar-spider}, it can be seen that in case study 1, PTQ and QAT both have equivalent overall compression successes, but BNN provides the best overall compression success with different trade offs compared to PTQ and QAT. Which compression techniques to be chosen is then down to the practitioners requirements.
In case study 2 the quantization to TFlite INT8  had the best overall compression success. 

\begin{figure}[]
    \centering
    \begin{subfigure}{0.49\textwidth}
        \includegraphics[width=1\textwidth]{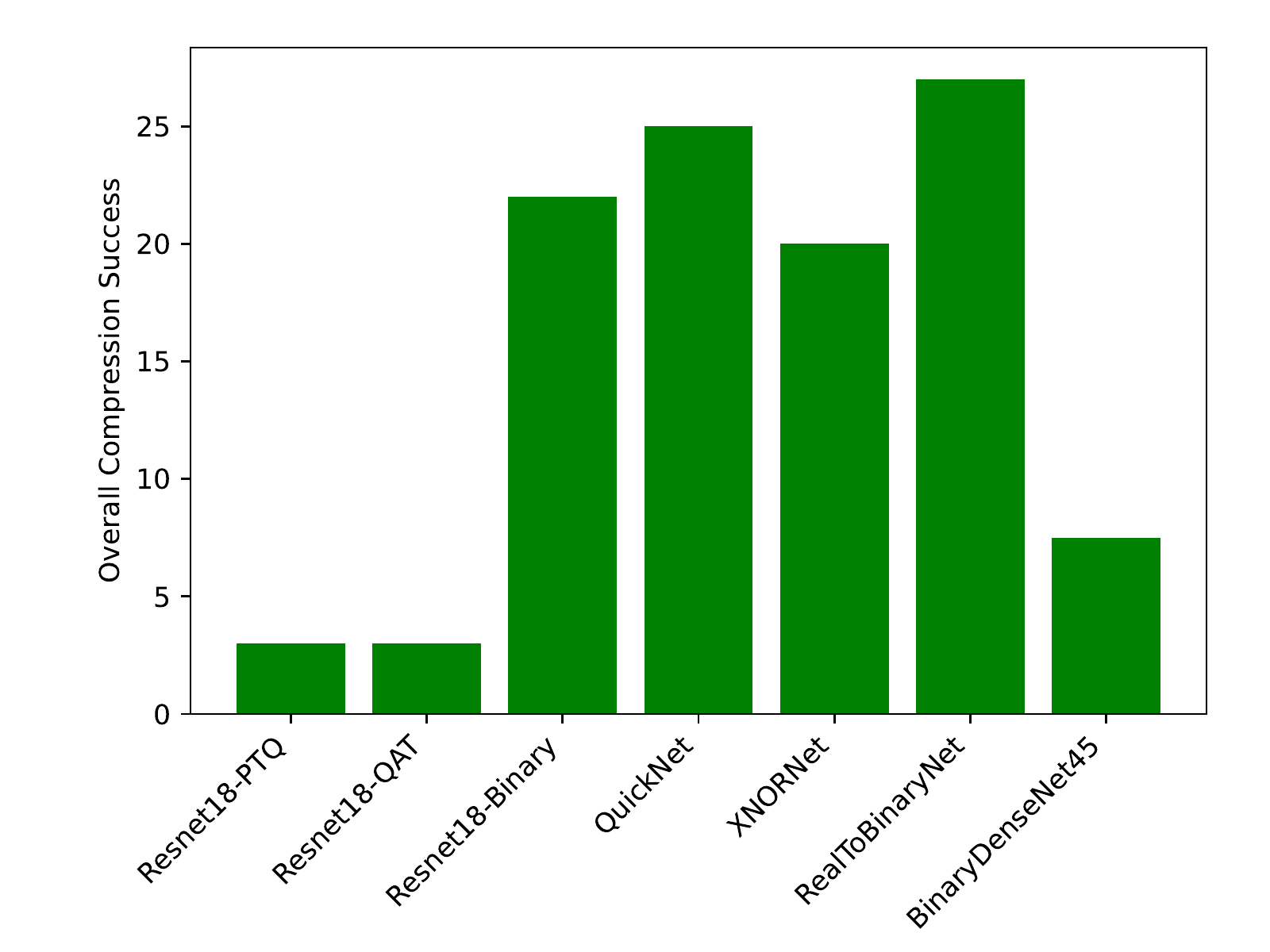}
        \caption{}
        \label{fig:various-a}
    \end{subfigure}
    \begin{subfigure}{0.49\textwidth}
        \includegraphics[width=1\textwidth]{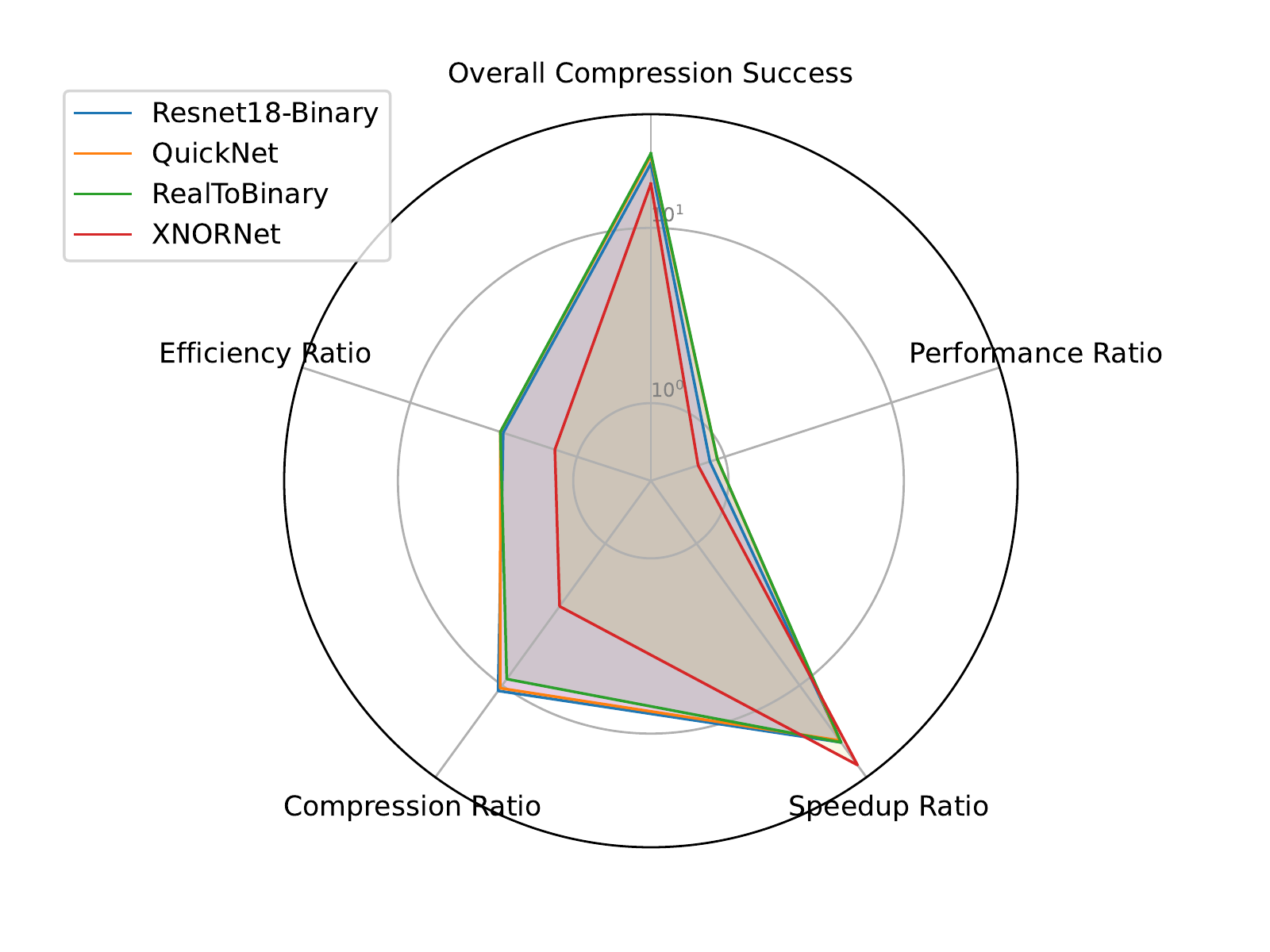}
        \caption{}
         \label{fig:various-b}
    \end{subfigure}
    \caption{(a) Bar plot showing Overall Compression Success for the different compressed neural networks in case study 3. (b) Radar plot for compressed models with top four OCS scores in case study 3.}
    \label{fig:various}
\end{figure}

Case study 3 shows a more a practical scenario for the usage of the OCS metric, where we are trying to compare between many compressed models. 
Using an initial assessment filter that shows the overall benefit of each compressed neural network can be more time efficient and easier to perceive, especially in commercial environments. 
In this case we are trying to find the most optimally compressed models for predicting Imagnet1k compared to our baseline (non-compressed Resnet18). 
There are seven models compared, two quantised using PyTorch and five binarized using LARQ. 
Figure~\ref{fig:various-a} compares between the different compressed models using the OCS metric. We can see that there are four compressed neural networks that have a close overall compression success to each other: Resnet18Binary, QuickNet, XNORNet, and RealToBinaryNet. These four top performing compressed models are plotted on a radar plot in Figure~\ref{fig:various-b}. 
The XNOR model seems to be the fastest but is the worst in all other improvement ratios. 
Excluding XNOR, the other filtered models are very close to each other in their polygons shapes. 
In this case we decide to prioritise the performance ratio, which makes the RealToBinary model the most efficient option for Imagnet1k out of our compared models.  

\begin{table}[H]
    \centering
    \begin{tabular}{| l || l | l | l | l | l | l | l | l | l |l|} 
        \hline
        Compression & Top-1 & Latency & MACs & CHATS & Disk & Params & CPU & RAM & Energy & Power \\
        Technique & Accuracy & (ms) &  $(\times 10^9)$ & $(\times 10^9)$ & Size & count & util & usage &(J) & (W) \\
         &  & & & & (MB) & $\times 10^6$& (\%) & (GB) & & \\
        \hline        
        None (FP32)        & 70.3 & 14 & 1.81         & 57.92 & 44.7 & 11.2 & 49.6 & 2.48 & 2.2 & 117.8\\
        PTQ (INT8)         & 70.1 & 7  & 1.81         & 14.48 & 11.3 & 11.2 & 49.1 & 2.52 & 0.8 & 111.6\\
        QAT (INT8)         & 69.6 & 7  & 1.81         & 14.48 & 11.3 & 11.2 & 49.1 & 2.51 & 0.8 & 111.6 \\
        GUP$_R$ (FP32)     & 50.8 & 15 & 1.81 (0.45)  & 57.92 (14.48) & 44.7 (11.2)& 11.2 (3.3) & 49.6& 2.51 & 1.8 & 121.8 \\
        GUP$_{L1}$ (FP32)  & 66.6 & 14 & 1.81 (0.45)  & 57.92 (14.48) & 44.7 (11.2)& 11.2 (3.3) & 49.6 & 2.46 & 1.7 &122.0\\
        BNN                & 58.3 & 74 & 1.81         & 1.81 & 4.0  & 11.2 & 35.1 & 1.03 & 4.3 & 58.1  \\
        \hline
    \end{tabular}
    \caption{Case Study 1: Summary of results for Resnet-18 inference on ImageNet1k running on PC.}
    \label{tab:resnet-imagnet1k}
\end{table}

\begin{table}[H]
    \centering
    \begin{tabular}{| l || l | l | l | l | l | l | l | l | l | l|} 
        \hline
        Compression & mAP & Latency & MACs & CHATS & Disk & Params & CPU & RAM & Energy & Power \\
        Technique &  & (ms) & $(\times 10^9)$& $(\times 10^9)$ & Size & count & util & usage &(J) & (W) \\
         &  & & & & (MB) & $(\times 10^6)$ & (\%) & (GB) & & \\
        \hline        
        None PyTorch (FP32)     & 56.6 & 3165 & 8.1 & 259.2  & 29.2 & 7.2 & 75.6 & 1.9  & 13.4 & 4.2 \\
        None TF (FP32)          & 56.5 & 1270 & 8.1 & 259.2 & 29.2 & 7.2 & 89.6 & 1.13 & 7.0 & 5.4 \\
        PTQ TFlite (FP16)  & 56.5 & 2288 & 8.1 & 129.6 & 14.6 & 7.2 & 25.1 & 0.49 & 8.9 & 3.8 \\
        PTQ TFlite (INT8)  & 54.7 & 1393 & 8.1 & 64.8 & 7.6  & 7.2 & 24.9 & 0.37 & 5.3 & 3.7 \\
        \hline
    \end{tabular}
    \caption{Case Study 2: Summary of results for YOLOv5s inference on COCO running on RasPi 4.}
    \label{tab:Yolov5s-COCO}
\end{table}

\begin{table}[H]
    \centering
    \begin{tabular}{| l || l | l | l | l | l | l | l | l | l |l|} 
        \hline
        Compressed Neural & Top-1 & Latency & MACs & CHATS & Disk & Params & CPU & RAM & Energy & Power \\
        Network  & Accuracy & (ms) &  $(\times 10^9)$ & $(\times 10^9)$ & Size & count & util & usage &(J) & (W) \\
         &  & & & & (MB) & $\times 10^6$& (\%) & (GB) & & \\
        \hline        
        Resnet-18 (FP32) - Baseline       & 70.3  & 14 & 1.81  & 57.92 & 44.7 & 11.2 & 49.6 & 2.48 & 2.2 & 117.8\\
        Resnet-18 - PTQ (INT8)            & 70.1  & 7  & 1.81  & 14.48 & 11.3 & 11.2 & 49.1 & 2.52 & 0.8 & 111.6\\
        Resnet-18 - QAT (INT8)            & 69.6  & 7  & 1.81  & 14.48 & 11.3 & 11.2 & 49.1 & 2.51 & 0.8 & 111.6 \\
        Resnet18 - Binary                   & 58.3  & 74 & 1.81  &  1.81 & 4.0  & 11.2 & 35.1 & 1.03 & 4.3 & 58.1  \\
        QuickNet~\cite{Bannink2020}                          & 63.4  & 85 & 1.88  &  1.88 & 4.2  & 13.2 & 34.3 & 1.09 & 4.7 & 55.8  \\
        XNORNet~\cite{Rastegari2016}                              & 44.9  &170 & 1.14  &  1.14 & 22.8 & 62.4 & 70.0 & 1.41 & 15.9& 93.5  \\
        RealToBinaryNet~\cite{Martinez2020Training}                      & 65.0  & 90 & 1.81  &  1.81 & 5.1  & 12.0 & 34.5 & 0.92 & 4.9 & 54.9  \\
        BinaryDenseNet~\cite{Bethge2019}                     & 64.6  &155 & 6.67  &  6.67 & 7.4  & 13.9 & 52.3 & 0.75 & 7.6 & 49.0      \\
        \hline
    \end{tabular}
    \caption{Case Study 3: Summary of results of various quantised neural networks inference on ImageNet1k running on PC.}
    \label{tab:various-BNN-Imagnet1k}
\end{table}

\section{Conclusions and Future Works} \label{sec:Conclusions}
In this paper, we have provided a review of evaluation metrics for neural network compression, with the aim of standardising evaluation of such techniques. 
The metrics are grouped into five categories: Accuracy, Speed, Size, Energy, and Combined Measures.
We identified lacking metrics that are needed to achieve more effective comparisons between compression techniques. 
Two novel evaluation metrics were contributed to cover identified gaps: 1) Compression and Hardware Agnostic Theoretical Speed (CHATS) and 2) Overall Compression Success (OCS). 
Reviewed and contributed metrics were implemented and provided in a library named NetZIP, available publicly.

We demonstrated the usage of reviewed and contributed metrics in three different case studies focusing on object classification, and object detection using two hardware platforms, a regular PC and a Raspberry Pi 4.
Throughout the development of this work, we used quantization and pruning as compression strategies. 
Future work includes expanding our evaluations in NetZIP to include other compression strategies (e.g. Knowledge Distillation, Tensor Decomposition).
Expanding to natural language processing (NLP) applications is also another area to further investigate.

Carrying out this work, we have identified other interesting areas of research that seem to not have attracted a lot research interest. 
Currently, pruning techniques only zero parameters, but, pruned parameters are not removed from the model automatically, as removal of parameters from an architecture can result in errors.
There is a need for developing model-agnostic techniques for the removal of pruned parameters.

Furthermore, there is scope for development of new metrics focused on  verification transferability, pre- and post-compression, to quantify functional equivalence.
Finally, there is need for doing more feasibility studies for utilisation of compressed models on edge devices.

\section{Acknowledgements}
The work has been funded by the Thales Bristol Partnership in Hybrid Autonomous Systems Engineering (T-B PHASE).
This paper is based upon work from the COST Action no. CA19135 ``Connecting Education and Research Communities for an Innovative Resource Aware Society (CERCIRAS)", supported by the European Commission through the European Cooperation in Science and Technology (COST) Association.

\printbibliography


\end{document}